\documentclass{article}

    \usepackage[preprint]{neurips_data_2024}

\usepackage[utf8]{inputenc} 
\usepackage[T1]{fontenc}    
\usepackage{hyperref}       
\usepackage{url}            
\usepackage{booktabs}       
\usepackage{amsfonts}       
\usepackage{nicefrac}       
\usepackage{microtype}      
\usepackage{xcolor}         

\usepackage{graphicx}
\usepackage{colortbl}
\usepackage{natbib}
\setcitestyle{numbers,square}

\usepackage{pifont}       
\usepackage{bbding}       
\usepackage{fontawesome}
\usepackage{subcaption}
\usepackage{multirow}

\begin{document}

\title{LaMOT: Language-Guided Multi-Object Tracking}

%

\author{%
   Yunhao Li$^{1,2}$,  \ \ \ \ Xiaoqiong Liu$^{3}$,  \ \ \ \ Luke Liu$^{4}$, \ \ \ \ Heng Fan$^{3,\dagger}$, \ \ \ \ Libo Zhang$^{2,\dagger,*}$ \\
   $^1$Institute of Software Chinese Academy of Science \\
   $^2$University of Chinese Academy of Science \\
   $^3$University of North Texas \ \ \ \ $^4$Intern at University of North Texas \\
  \texttt{liyunhao23@mails.ucas.ac.cn, heng.fan@unt.edu, libo@iscas.ac.cn} \\
  $\dagger$Equal Advising \;\;\; $*$Corresponding Author\\
}






\maketitle

\begin{abstract}

Vision-Language MOT is a crucial tracking problem and has drawn increasing attention recently. It aims to track objects based on human language commands, replacing the traditional use of templates or pre-set information from training sets in conventional tracking tasks. Despite various efforts, a key challenge lies in the lack of a clear understanding of \textit{why language is used for tracking}, which hinders further development in this field. In this paper, we address this challenge by introducing \textbf{Language-Guided MOT}, a \textit{unified} task framework, along with a corresponding large-scale benchmark, termed \textbf{LaMOT}, which encompasses diverse scenarios and language descriptions. Specially, LaMOT comprises 1,660 sequences from 4 different datasets and aims to unify various Vision-Language MOT tasks while providing a standardized evaluation platform. To ensure high-quality annotations, we manually assign appropriate descriptive texts to each target in every video and conduct careful inspection and correction. To the \textit{best} of our knowledge, LaMOT is the \textit{first} benchmark dedicated to Language-Guided MOT. Additionally, we propose a simple yet effective tracker, termed \textbf{LaMOTer}. By establishing a unified task framework, providing challenging benchmarks, and offering insights for future algorithm design and evaluation, we expect to contribute to the advancement of research in Vision-Language MOT. We will release the data at \url{https://github.com/Nathan-Li123/LaMOT}.

\end{abstract}

\section{Introduction}

Multi-Object Tracking (MOT) is an important task in computer vision, which has garnered significant attention, leading to the emergence of various innovative approaches \cite{ye2022joint, yan2022towards, gao2023memotr, cao2023observation, zeng2022motr}. Recently, there has been a marked surge of interest within the MOT community towards integrating natural language processing into MOT approaches, termed \textit{Vision-Language MOT}. This integration aims to track areas or targets of interest based on human language instructions. In particular, several approaches and benchmarks (\textit{e.g.}, \cite{wu2023referring, nguyen2024type, li2023ovtrack}) have been proposed, significantly facilitating related research endeavors and advancements on this topic. However, despite these efforts, we argue there is still a misunderstanding of a crucial question: \textbf{\textit{why language is used for tracking?}} In this paper, we summarize the answer as two key words: \textbf{\textit{flexibility}} and \textbf{\textit{generality}}.

\begin{figure*}[!t]
 \centering
 \includegraphics[width=1.0\linewidth]{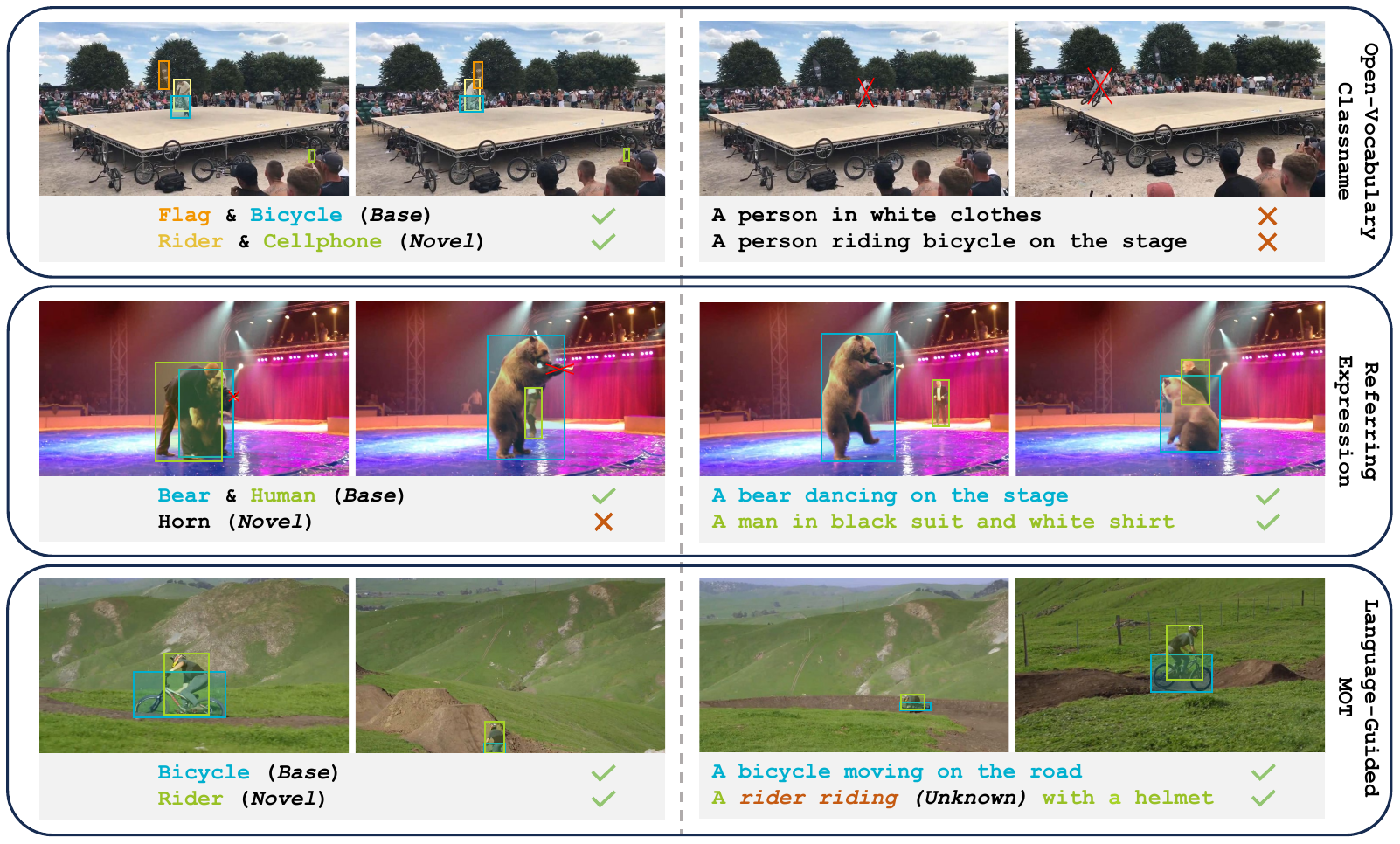}
 \caption{The visualization of two main settings of Vision-Language MOT, \textit{i.e.}, \textit{open-vocabulary classname tracking} and \textit{referring expression tracking}. Moreover, Language-Guided MOT enables tracking with any form of language while possessing the ability to recognize open-vocabulary terms.}
 \label{fig:intro}
 \vspace{-5mm}
\end{figure*}

Vision-Language MOT tasks can be typically classified into two settings: \textit{open-vocabulary classname tracking} and \textit{referring expression tracking} (see Fig.~\ref{fig:intro}). Although these definitions seem reasonable, they inadvertently restrict the \textbf{\textit{flexibility}} of natural language. Open-vocabulary classname tracking approaches focus on empowering models to track \textit{unknown} categories, but they are constrained by the conventional MOT category concept, unable to recognize more complex yet practical language descriptions. On the other hand, referring expression tracking methods aim to ensure that models comprehend \textit{closed-set} language descriptions, but they struggle when facing open-vocabulary contexts as analyzed in~\cite{wu2023referring}. To this end, we introduce \textbf{Language-Guided MOT}, a \textit{unified} task framework for Vision-Language MOT. As shown in Fig.~\ref{fig:intro}, Language-Guided MOT combines the advantages of both settings, enabling tracking with any form of language while possessing the ability to recognize open-vocabulary terms. We note that the open-vocabulary capability required by Language-Guided MOT is reflected in the entire vocabulary used in language descriptions, rather than being limited to category names. This maximizes the \textit{flexibility} of using natural language in MOT.





Besides task definition, Vision-Language MOT benchmarks also face severe challenges. First, following existing tasks definitions, vision-langauge benchmarks~\cite{wu2023referring, li2023ovtrack} tend to revolve around only one challenge factor: they either prioritize incorporating \textit{open-set categories} or lean towards utilizing \textit{closed-set descriptions}. This weaken the challenges posed by real-world Vision-Language MOT where arbitrary challenges may exist and severely limit the \textit{\textbf{flexibility}} of natural language. Second, a crucial point is largely overlooked in previous works: video \textit{scenarios}. Conventional tracking tasks typically rely on templates or predefined information from the training set to determine the targets to be tracked. This directly leads to a significant degradation in model performance when there are noticeable changes in video scenarios, as it is very challenging for the model to gain the ability of extracting target information from different viewpoints. However, owing to its inherited \textit{\textbf{generality}}, natural language can mitigates this issue to a large extent from a \textit{multimodal} angle. Despite the existence of various Vision-Language MOT benchmarks covering different scenarios, they focus on only one or at most two video scenarios within each individual dataset.

To foster the study of Language-Guided MOT, we propose a large-scale benchmark, termed \textbf{LaMOT}. Specifically, LaMOT comprises 1,660 sequences, 1.67M frames, and over 18.9K target trajectories (see Tab.~\ref{tab:dataset-comparison-01}). These video sequences are sourced from four datasets, \textit{i.e.}, MOT17~\cite{milan2016mot16}, TAO~\cite{dave2020tao}, VisDrone2019~\cite{wen2019visdrone}, and SportsMOT~\cite{cui2023sportsmot}. They encompass five different scenarios including \textit{surveillance}, \textit{autonomous driving}, \textit{sports broadcasting}, \textit{drone}, and \textit{daily life}. In addition, we meticulously design appropriate descriptive sentences for each trajectory, ensuring the accuracy of the annotations through careful inspection and refinement. To our knowledge, LaMOT is the \textit{largest} and \textit{most challenging} publicly available Vision-Language MOT benchmark to date and the \textit{first} benchmark dedicated to Language-Guided MOT. By releasing LaMOT, we aim to provide a dedicated platform for advancing in the research on Language-Guided MOT. 

Furthermore, to better facilitate research in this field, we propose a simple yet effective baseline \textbf{LaMOTer}. Specially, LaMOTer combine GroundingDINO's~\cite{liu2023grounding} text-based detection capabilities with OC-SORT's~\cite{cao2023observation} robust tracking and matching abilities. We conduct experiments on LaMOT using LaMOTer and a series of established trackers. Besides the analysis on overall performance, we independently study the difficulty between different video scenarios. We further conduct an in-depth analysis of the evaluation results and hope these evaluations and analyses can offer baselines for future research in Language-Guided MOT, providing guidance for tracking algorithm design.

In summary, our main contributions are as follows: 
\ding{168} We introduce a novel task, termed \textbf{Language-Guided MOT} to unify various Vision-Language MOT tasks that share similar underlying principles. 
\ding{169} We propose \textbf{LaMOT}, which to our knowledge is the largest, most standardized, and most challenging benchmark in the relevant field. 
\ding{170} We propose \textbf{LaMOTer}, a simple yet effective tracker to facilitate future research. 
\ding{171} We conduct experiments and in-depth analysis for the evaluations of the proposed approach and benchmark, providing guidance for future algorithm design.

\section{Related Works}

\subsection{Multi-Object Tracking}

Multi-Object Tracking (MOT) involves detecting and tracking multiple moving objects in video sequences while ensuring consistent identities across frames. It's vital for applications like video surveillance, autonomous driving, and sports analysis. 

Benchmarks have always been pivotal in advancing the development of MOT. One of the earliest benchmarks, PETS2009~\cite{ferryman2009pets2009}, focuses on multi-pedestrian tracking. The MOT Challenge~\cite{milan2016mot16, dendorfer2020mot20}, featuring more crowded videos, has significantly propelled MOT forward. ImageNet-Vid~\cite{deng2009imagenet} provides trajectory annotations for 30 categories across more than 1,000 videos, whereas TAO~\cite{dave2020tao} expands this to include 833 object classes for general multi-object tracking. For specialized areas like dancing and sports, DanceTrack~\cite{sun2022dancetrack} and SportsMOT~\cite{cui2023sportsmot} were developed to track dancers and players. And for autonomous driving, KITTI~\cite{geiger2013vision} and BDD100K~\cite{yu2020bdd100k} were specifically created for object tracking. AnimalTrack~\cite{zhang2023animaltrack} targets the tracking of various animals in natural environments. Additionally, VisDrone~\cite{wen2019visdrone} provides benchmarks for tracking objects using drones.

MOT algorithms have made significant strides in recent years. A widely adopted approach is the \emph{Tracking-by-Detection} paradigm, where objects are detected first and then associated across frames. This method underpins many notable techniques~\cite{wojke2017simple, du2023strongsort, cao2023observation, maggiolino2023deep, zhang2022bytetrack}. Improvements in detection accuracy and matching effectiveness is crucial for enhancing performance in these methods. Another common approach is the \emph{Joint-Tracking-and-Detection} paradigm~\cite{ye2022joint, yan2022towards, zhou2020tracking, han2020unicorn, yan2023universal}, which combines tracking and detection into a single, end-to-end process. Recently, the use of Transformers~\cite{vaswani2017attention} in MOT has led to remarkable improvements, surpassing previous trackers~\cite{sun2020transtrack, meinhardt2022trackformer, zeng2022motr, chu2023transmot, zhang2023motrv2, gao2023memotr, zhou2022global}.

\subsection{Vision-Language MOT Benchmarks}

Vision-Language MOT integrates computer vision and NLP to track multiple objects in videos using textual descriptions. This approach leverages visual data and language cues to enhance tracking accuracy and flexibility, enabling more effective object tracking in dynamic environments.

Benchmarks are important for the development of Vision-Language MOT. In recent years, many benchmarks have been proposed. Ref-YTVIS~\cite{seo2020urvos}, building upon Youtube-VOS, introduces text annotations in two forms: \textit{full-video} and \textit{first-frame}. These additions significantly contribute to Vision-Language MOT tasks as well as segmentation tasks. TAO~\cite{li2023ovtrack} largely adheres to the taxonomy established by LVIS~\cite{gupta2019lvis}, which categorizes classes based on their occurrence as frequent, common, and rare. OV-TAO, building upon TAO, follows open-vocabulary detection literature by dividing categories into \textit{base} and \textit{novel} classes, fostering the development of Open-Vocabulary MOT. Refer-KITTI~\cite{wu2023referring}, an extension of the KITTI dataset~\cite{geiger2013vision}, focuses on using referential expressions in traffic scenes for MOT. Grounded Multiple Object Tracking (GroOT)~\cite{nguyen2024type} is a recently introduced dataset featuring videos of various objects along with detailed textual captions describing their appearance and actions. \textbf{\textit{Different from}} existing datasets, our proposed LaMOT combines various language settings, tracking scenarios, and shooting perspectives, and has undergone standardized adjustments to the language texts. It is the \textit{largest} and \textit{most challenging} dataset in the field to date.

\begin{table*}[t]
\centering
\renewcommand{\arraystretch}{1}
\tabcolsep=2mm
\caption{Comparison between LaMOT and several related datasets. \# denotes the number of the corresponding item, ``-'' denotes that the corresponding item is not provided and ``n/a'' denotes that data is not officially available. ``\#Scenes'' represents the number of scenarios in the dataset, ``OV'' and ``RE'' respectively denote open-vocabulary classname and referring expression settings. \textbf{Bold} numbers are the best number in each sub-block, while \colorbox[rgb]{1,0.8,0.6}{highlighted} numbers are the best across all.}
\resizebox{1.0\textwidth}{!}{
\begin{tabular}{l|cccccccc}
    \toprule[1.2pt]
    \textbf{Datasets} & \textbf{\#Videos} & \textbf{\#Frames} & \textbf{\#Boxes} & \textbf{\#Tracks} & \textbf{\#Scenes} & \textbf{\#Words} & \textbf{OV} & \textbf{RE} \\ 
    \midrule[0.8pt]
    \textbf{MOT17}~\cite{milan2016mot16} & 14 & 11.2K & \textbf{300K} & 1.3K & 1 & - & - & - \\
    \textbf{MOT20}~\cite{dendorfer2020mot20} & 8 & 13.41K & 2.1M & 3.83K & 1 & - & - & - \\
    \textbf{TAO}~\cite{dave2020tao} & \textbf{1.5K} & \cellcolor[rgb]{1,0.8,0.6}{\textbf{2.2M}} & 170K & \textbf{8.1K} & \textbf{2} & - & - & - \\
    \textbf{DanceTrack}~\cite{sun2022dancetrack} & 100 & 106K & n/a & 990 & 1 & - & - & - \\
    \midrule[0.8pt]
    \textbf{OV-TAO}~\cite{li2023ovtrack} & 1.5K & \cellcolor[rgb]{1,0.8,0.6}{\textbf{2.2M}} & 170K & 8.1K & 2 & 170K & $\checkmark$ & -  \\
    \textbf{Ref-KITTI}~\cite{wu2023referring} & 18 & 6.65K & 34K & 660 & 1 & 3.7K & - & $\checkmark$ \\
    \textbf{GroOT}~\cite{nguyen2024type} & 1,515 & 1.59M & 545.9K & 13.3K & 1 & 256K & - & $\checkmark$ \\
    \textbf{LaMOT(Ours)} & \cellcolor[rgb]{1,0.8,0.6}{\textbf{1,660}} & 1.67M & \cellcolor[rgb]{1,0.8,0.6}{\textbf{2.44M}} & \cellcolor[rgb]{1,0.8,0.6}{\textbf{18.9K}} & \cellcolor[rgb]{1,0.8,0.6}\textbf{5} & \cellcolor[rgb]{1,0.8,0.6}\textbf{2.62M} & $\checkmark$ & $\checkmark$ \\
    \bottomrule[1.2pt]
\end{tabular}
\label{tab:dataset-comparison-01}
\vspace{-5mm}
}	
\end{table*}

\section{The Proposed LaMOT}

\subsection{Design Principle}

We propose LaMOT to provide a large-scale platform for Language-Guided MOT and offer a more challenging, yet standardized, testbed for evaluating vision-language trackers in a practical manner. To this end, We follow four principles in constructing our LaMOT:

(1)  \textit{Dedicated benchmark}. One major motivation behind LaMOT is to provide a dedicated benchmark for Language-Guided MOT. Given that the substantial training data required for deep learning models, we aim to establish a platform with at least 1,500 sequences and 1.5 million frames.
    
(2)  \textit{Diverse scenarios}. The variety of video scenarios is often overlooked in current datasets, yet it is crucial for developing a general system. To provide a diverse platform for Language-Guided MOT, we will incorporate five types of video sequences with different scenarios in LaMOT, \textit{i.e.}, \textit{surveillance}, \textit{autonomous driving}, \textit{sports broadcasting}, \textit{drone}, and \textit{daily life}.

(3)  \textit{High-quality annotations}. High-quality annotations are crucial for establishing a benchmark for both training and assessing models. In LaMOT, we meticulously examine each sequence and manually craft appropriate descriptive texts for each trajectory to ensure high-quality and standardized annotations. This process involves multiple rounds of inspection and refinement.

(4)  \textit{Varied trajectory density}. Current Vision-Language MOT benchmarks tend to have a relatively low average trajectory count per video, yet high-density scenes being common in real-world applications. Therefore, we hope to include a wide range of trajectory densities in LaMOT. We expect each sequence to contain between 1 and more than 300 trajectories.

\subsection{Data Collection}

LaMOT focuses on establishing a large-scale dataset that unifies both vision and language aspects. To achieve this goal, LaMOT requires video sequences with rich diversity in scenarios, video viewpoints, and target categories, significantly exceeding the demands of existing benchmarks. We initiate benchmark construction by selecting five common scenarios, including \textit{surveillance}, \textit{autonomous driving}, \textit{sports broadcasting}, \textit{drone}, and \textit{daily life}. After determining the required scenarios, we survey existing object tracking datasets and ultimately select four: VisDrone2019~\cite{wen2019visdrone}, which provides video sequences of the drone scenario, SportsMOT~\cite{cui2023sportsmot}, which offers sequences of the sports broadcasting scenario, MOT17~\cite{milan2016mot16}, which provides sequences of the pedestrian surveillance scenario, and TAO~\cite{dave2020tao}, which offers video sequences of both the autonomous driving and daily life scenario, along with a wide variety of target categories.

Eventually, we compile a large-scale dataset by gathering 1,660 sequences with 1.67M frames from four distinctive datasets. The average length of sequences in LaMOT is 1,008 frames, with the longest sequence containing 2,341 frames and the shortest one consisting of 58 frames. On average, each sequence contains 11.4 trajectories. The sparsest sequence contains only 1 target, while the densest sequence consists of more than 500 trajectories.

\subsection{Data Annotation}

In order to offer high-quality annotations for LaMOT, we manually annotate each object in every sequence with appropriate descriptions. Specifically, we observe the entire video and annotate the targets based on their \textit{appearance}, \textit{position}, and \textit{actions}. Notably, a single trajectory may be associated with multiple descriptions, and a single description may be relevant to multiple trajectories. 

While this strategy generally works well, there are exceptions. Describing certain states of a target, such as short-term states or variable attributes like a \textit{jumping} person, can be confusing. It is impractical to track some people while they are jumping and then stop tracking when they land. However, consistent state information may still provide valid descriptions for a target and thus should not be ignored. Therefore, during the annotation, we focus only on attributes that remain consistent throughout at least the vast majority of the video.

The most significant effort in constructing a large-scale dataset lies in manual labeling, double-checking, and error correction. To ensure high-quality annotations in LaMOT, we employ a multi-round strategy. Initially, volunteers familiar with the tracking domain and our annotation principles conduct the first round of labeling. Subsequently, experts review these initial annotations, and any issues are returned to the labeling team for revision. We repeat this process, facilitating communication between the labeling team and experts, until both parties are satisfied with all annotations.

\begin{figure*}[!t]
 \centering
 \includegraphics[width=\linewidth]{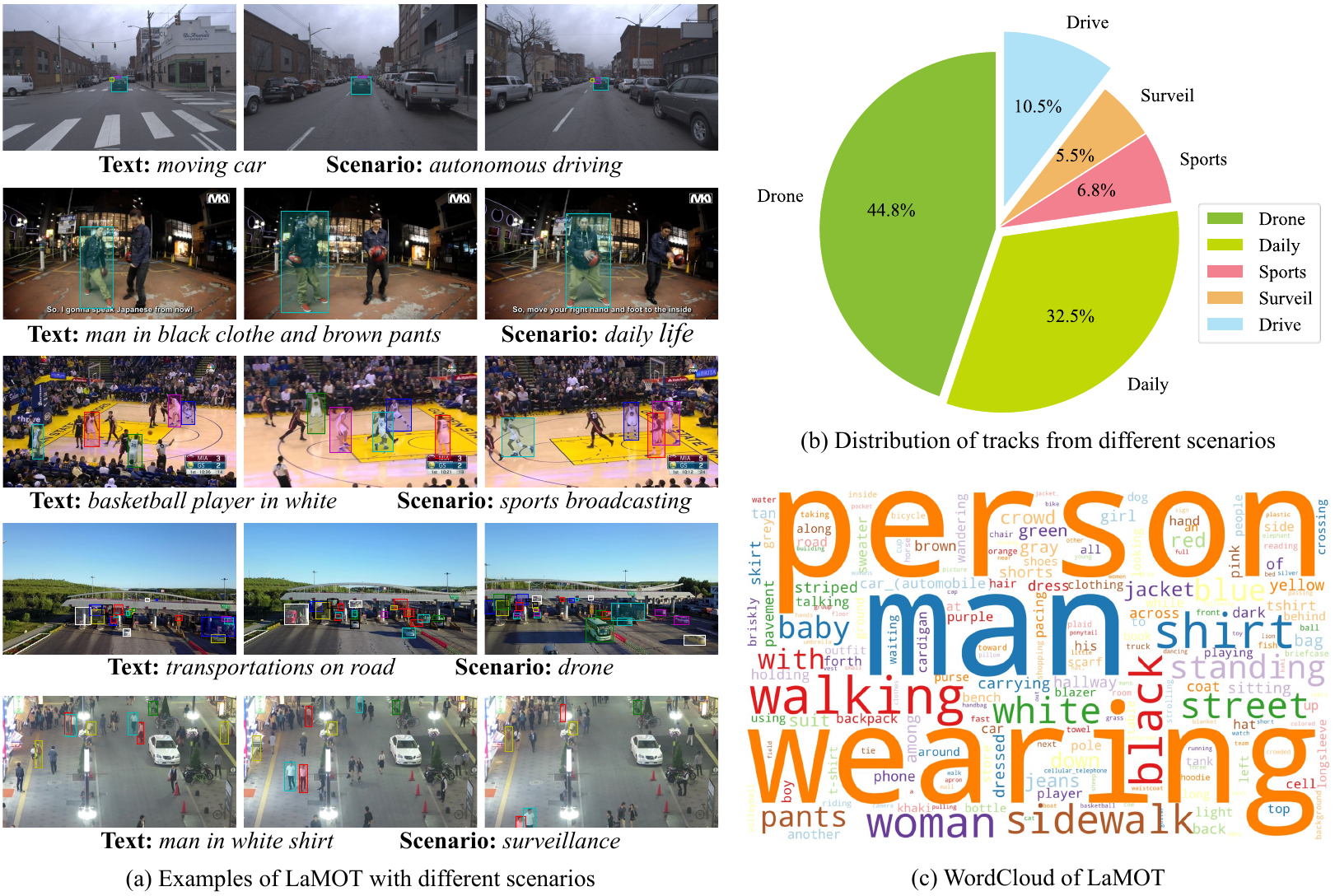}
 \caption{(a) Representative examples from LaMOT, including sequences from five different scenarios. (b) Distribution of track counts from different scenarios, highlighting two scenarios from the same dataset~\cite{dave2020tao}. (c) Word cloud of the vocabulary used in LaMOT.}
 \label{fig:examples}
 \vspace{-5mm}
\end{figure*}

\noindent
\textbf{Statistics of annotations.} We compare LaMOT with several related datasets (see Tab.~\ref{tab:dataset-comparison-01}). It is evident that LaMOT enjoys satisfactory data scale, annotation quantity, and diversity of scenarios. To further demonstrate LaMOT, we provide a more detailed comparison between LaMOT, Ref-KITTI, and GroOT. As shown in Tab.~\ref{tab:dataset-comparison-02}, notably, LaMOT offers more comprehensive annotations than existing datasets. Besides quantity and quality, LaMOT differs from previous datasets in two aspects. First, it includes sequences with high trajectory densities, featuring up to 500 tracks in a single video. Second, LaMOT's open vocabulary ability is reflected not only in category names but also in the vocabulary used in descriptions. Nearly \textit{one-third} of the total vocabulary in LaMOT consists of words that do not appear in the training set (see Tab.~\ref{tab:dataset-comparison-02}), demonstrating its robustness in handling diverse and unseen vocabulary. In addition, we showcase example sequences from LaMOT in Fig.~\ref{fig:examples}(a), track counts distribution of different scenarios in Fig.~\ref{fig:examples}(b), and the wordcloud of LaMOT in Fig.~\ref{fig:examples}(c). 

\vspace{-3mm}
\begin{table*}[t]
\centering
\renewcommand{\arraystretch}{1}
\tabcolsep=1.5mm
\caption{Detailed Comparison between LaMOT, Ref-KITTI~\cite{wu2023referring}, and GroOT~\cite{nguyen2024type}. ``Vocab.'' represents the total number of vocabulary used in the data split. Some item of the GroOT dataset is marked as n/a (not available) because it lacks publicly available annotations for the complete test set. }
\resizebox{1.0\textwidth}{!}{
\begin{tabular}{ll|cccccccc}
    \toprule[1.2pt]
    \multicolumn{2}{c|}{\multirow{2}{*}{\textbf{Datasets}}} & \multirow{2}{*}{\textbf{\#Videos}} & \multirow{2}{*}{\textbf{\#Frames}} & \multirow{2}{*}{\textbf{\#Boxes}} & \multirow{2}{*}{\textbf{\#Tracks}} & \textbf{Min.} & \textbf{Avg.} & \textbf{Max.} & \multirow{2}{*}{\textbf{Vocab.}}  \\ 
    \multicolumn{2}{c|}{} & & & & & \textbf{Tracks} & \textbf{Tracks} & \textbf{Tracks} & \\
    \midrule[0.8pt]
    \multirow{3}{*}{\textbf{Ref-KITTI}~\cite{wu2023referring}} & train & 15 & 5.64K & 28.0K & 526 & 3 & 35.1 & 113 & 38 \\
    & test & 3 & 1.01K & 6.0K & 134 & 33 & 44.7 & 57 & 44 \\
    \cmidrule(){2-10}
    & \textbf{total} & 18 & 6.65K & 34.0K & 660 & 3 & 36.7 & 113 & 50 \\
    \midrule[0.8pt]
    \multirow{3}{*}{\textbf{GroOT}~\cite{nguyen2024type}} & train & 507 & 539.4K & 259.3K & 3.2K & 1 & 6.3 & 133 & 962 \\
    & test & 1008 & 1.05M & 286.6K & 10.1K & n/a & 10.0 & n/a & n/a \\
    \cmidrule(){2-10}
    & \textbf{total} & 1,515 & 1.59M & 545.9K & 13.3K & 1 & 8.8 & n/a & n/a \\
    \midrule[0.8pt]
    \multirow{3}{*}{\textbf{LaMOT(Ours)}} & train & 608 & 592.1K & 1.74M & 11.7K & 1 & 19.4 & 500 & 992 \\
    & test & 1,052 & 1.08M & 700.3K & 7.15K & 1 & 6.8 & 184 & 675 \\
    \cmidrule(){2-10}
    & \textbf{total} & 1,660 & 1.67M & 2.44M & 18.9K & 1 & 11.4 & 500 & 1312 \\
    \bottomrule[1.2pt]
\end{tabular}
\label{tab:dataset-comparison-02}
\vspace{-0.5em}
}	
\end{table*}

\subsection{Dataset Split and Evaluation Metric}

\textbf{Dataset Split.} LaMOT is build upon four existing MOT datasets. Therefore, for the splitting of training and test sets, we \textit{mainly} follow the original settings of these datasets. For MOT17, since annotations for its test set are not available, we use 2 videos from its training set for testing. Specifically, the training set comprises 608 sequences with 592.1K frames, while the test set consists of 1,052 videos with 1.08M frames. More details are demonstrated in Tab.~\ref{tab:dataset-comparison-02}.

\textbf{Evaluation Metric.} As a unified benchmark, we do not further partition the categories into \textit{base} and \textit{novel} classes as done in ~\cite{li2023ovtrack}. In fact, we believe the concept of category does not even exists in Language-Guided MOT. For evaluation, we follow ~\cite{wu2023referring} and employ higher order tracking accuracy (HOTA), association accuracy (AssA), detection accuracy (DetA), and localization accuracy (LocA) by following~\cite{luiten2021hota}, CLEAR metrics~\cite{bernardin2008evaluating} including multiple object tracking accuracy (MOTA), false positives (FP), false negatives (FN), and ID switches (IDs), and ID metrics~\cite{ristani2016performance} containing identification precision (IDP), identification recall (IDR) and related F1 score (IDF1). 

\section{Methodology}

\textbf{Overview.}  To encourage development of Language-Guided trackers, in this paper we propose LaMOTer, a \textit{simple} but \textit{effective} approach to achieve Language-Guided MOT. As illustrated in Fig.~\ref{fig:arch}, LaMOTer can be logically divided in two key parts: \textit{vision-language detection} and \textit{object tracking}. We explain them in Sec.~\ref{sec:detect} and Sec.~\ref{sec:track}, respectively.


\subsection{Vision-Language Detection} \label{sec:detect}

Current state-of-the-art vision-language trackers~\cite{li2023ovtrack, nguyen2024type, wu2023referring} are unable to handle both the comprehension of open-vocabulary category names and the understanding of arbitrary forms of descriptive text \textit{simultaneously}. To address this problem, we draw inspiration from GroundingDINO~\cite{liu2023grounding}, an advanced model for visual grounding designed to locate objects in images based on textual descriptions. GroundingDINO employs attention mechanisms to accurately identify and localize objects in images, making it highly suitable for tasks such as image captioning, and object detection. Most importantly, it meets both of our requirements: recognizing category names in an open-vocabulary format and understanding diverse forms of descriptive statements.

\begin{figure*}[!t]
 \centering
 \includegraphics[width=1.0\linewidth]{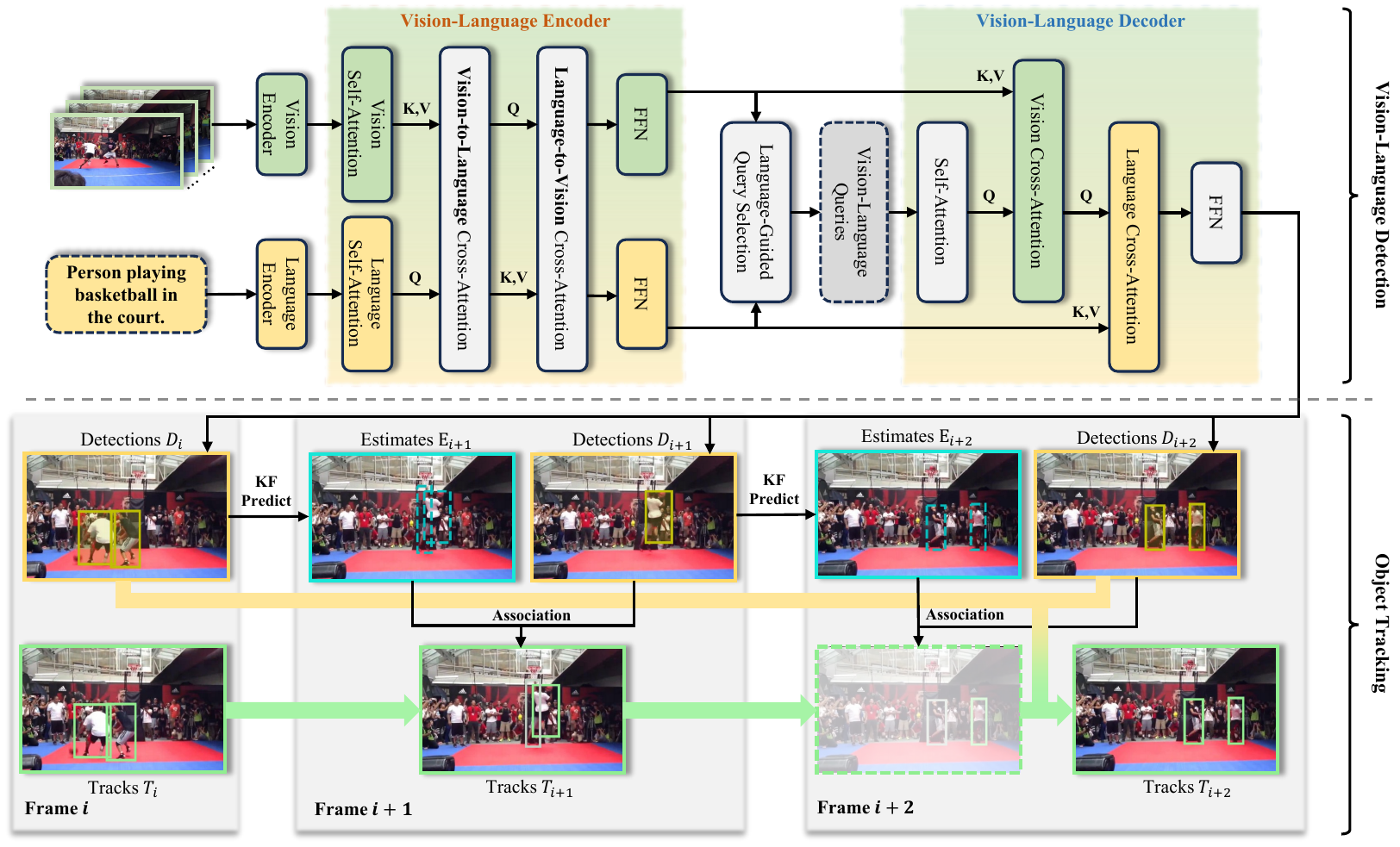}
 \caption{Illustration of our proposed LaMOTer, which contains two components of vision-language detection and object tracking. ``KF'' stands for Kalman Filter~\cite{kalman1960new}.}
 \label{fig:arch}
 \vspace{-5mm}
\end{figure*}

Give a video with $N$ frames and a input text, LaMOTer deal with them as $N$ pairs of \texttt{(Image, Text)}. For each pair, LaMOTer first extracts plain vision feature and plain language feature using a transformer-based vision encoder and a language encoder (We use BERT~\cite{devlin2018bert} in LaMOTer), respectively. Then, a Vision-Language Encoder enhances the two plain features through cross-fusion to obtain enhanced features (see Fig.~\ref{fig:arch}). Afterwards, LaMOTer uses a Language-Guided Query Selection module to select cross-modality queries, essentially using language to highlight important areas of the image. Lastly, LaMOTer decodes the cross-modality queries and the two enhanced features with a DETR-like~\cite{carion2020end} Vision-Language Decoder to produce the final detection outputs. 

\subsection{Object Tracking} \label{sec:track}

During our comprehensive review of existing works, we identify a significant limitation prevalent in current vision-language tracking approaches: these methods exhibit considerable difficulty in mitigating the influence of object detection when aligning targets with textual descriptions. Typically, current methods adopt one of two paradigms: either integrating image and text features prior to object detection or conducting object detection first and subsequently matching the detected object features with the corresponding textual descriptions. Although both paradigms offer their respective advantages, they share a common critical issue: the efficacy of multimodal matching is intrinsically linked to the quality of object detection. This linkage introduces a notable bias, as targets that are easier to detect naturally exhibit superior performance in multimodal matching tasks.

To address this bias, we employ a straightforward yet effective strategy: elevating the threshold for multimodal matching. While this approach allows us to identify more precise targets, it also inherently increases the risk of significant target loss. To mitigate this issue, we employ OC-SORT~\cite{cao2023observation} in the second phase of LaMOTer, which provides a robust solution.

OC-SORT mainly follows SORT~\cite{bewley2016simple}, utilizing a Kalman filter~\cite{kalman1960new} to predict motions. The Hungarian algorithm is then employed to associate detection boxes with predicted boxes based on Intersection over Union (IoU), enabling real-time tracking. But unlike standard SORT, OC-SORT effectively recovers lost targets mid-track through observation (see Fig~\ref{fig:arch}), thanks to two unique modules: \textit{Observation-Centric Re-Update} (ORU) and \textit{Observation-Centric Momentum} (OCM). ORU enhances the Kalman filter's ability to update the state of the target using observational data. This dynamic adjustment of the target's position, size, and orientation based on current frame observations allows for better adaptation to the target's motion and appearance changes. Concurrently, OCM fine-tunes the tracker’s velocity and direction by analyzing observational data, thereby improving the tracking of the target's motion trajectory. These enhancements enable OC-SORT to effectively recover lost targets during tracking, thereby addressing the problem of target loss that arises from increasing the multimodal matching threshold. Empirical experiments validate the efficacy of LaMOTer.

\begin{figure*}[!t]
 \centering
 \includegraphics[width=1.0\linewidth]{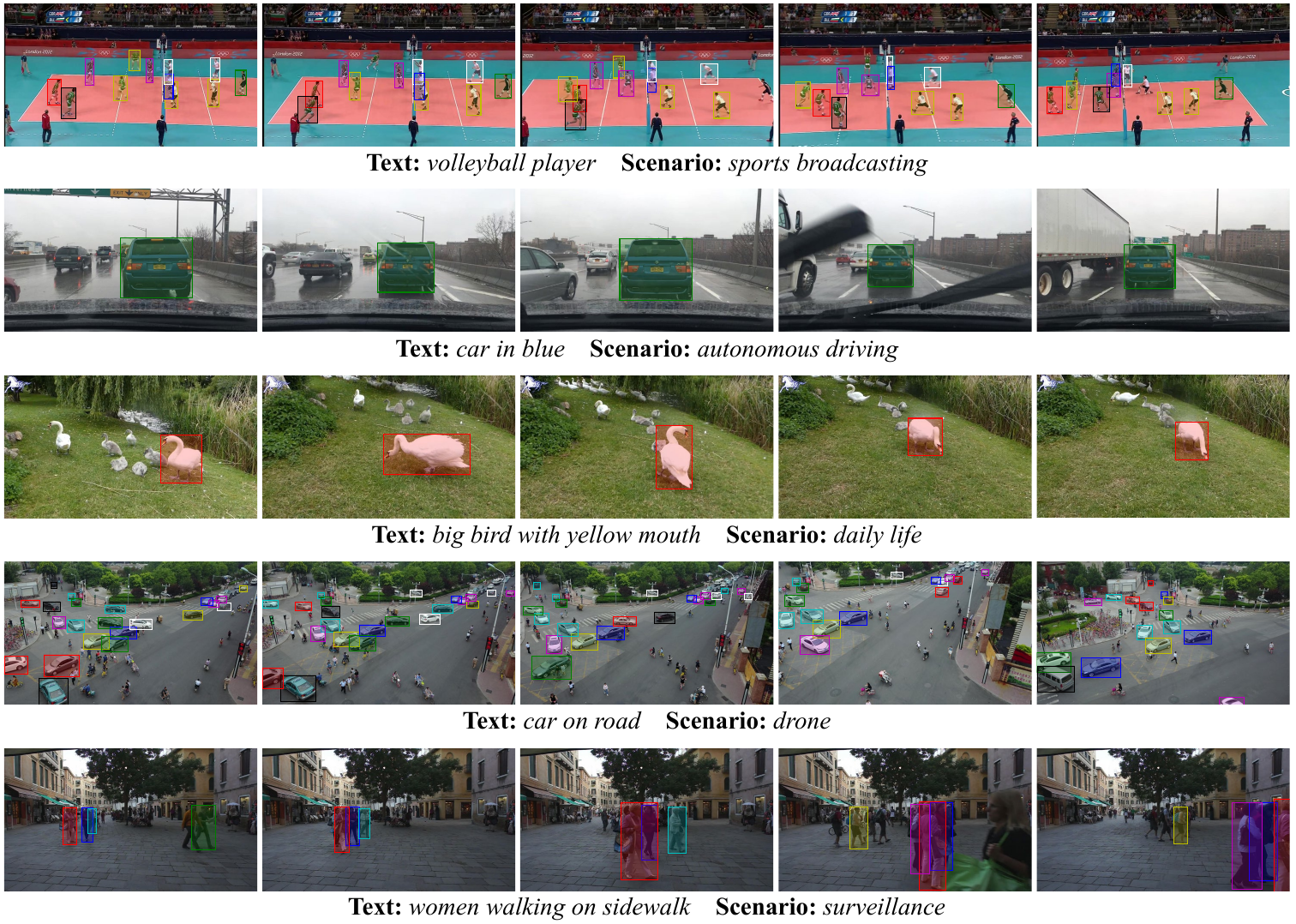}
 \caption{Qualitative results of LaMOTer on LaMOT. We observe that LaMOTer achieves satisfactory on all 5 scenarios. Each color represents a tracking trajectory. }
 \label{fig:results}
 \vspace{-5mm}
\end{figure*}

\section{Experiments}

\subsection{Experimental Setup}

\textbf{Experimental setup.} Since Language-Guided MOT is a novel unified task, there aren't existing models perfectly suit for it (In fact, this motivates the introduction of LaMOT and LaMOTer to foster research on Language-Guided MOT). We first compare LaMOTer with a series of established two-stage trackers, leveraging several state-of-the-art and classic models including SORT~\cite{bewley2016simple}, DeepSORT~\cite{wojke2017simple}, BYTETrack~\cite{zhang2022bytetrack}, StrongSORT~\cite{du2023strongsort}, and MOTRv2~\cite{zhang2023motrv2}. We further evaluate TransRMOT~\cite{wu2023referring}, which to our knowledge is the only publicly available approach capable for Language-Guide MOT. In addition, we analyze the difficulty of different scenarios.

\textbf{Implementation details.} We conduct our experiments using 4 Nvidia Tesla V100 GPUs with 32GB of VRAM. For the methods we devise, we configure the batch size to 4 and employ the AdamW optimizer with an initial learning rate of $5.0\times10^{-5}$. Throughout training, we discard tracked targets with scores below the threshold $\tau = 0.5$. For lost tracklets, we preserve them for 30 frames in anticipation of their reappearance. We utilize the original architectures of all selected approaches without any modifications and train them on our LaMOT dataset. 


\subsection{Overall Performance}

To evaluate the effectiveness of our proposed LaMOTer, we compare it with several established two-stage trackers. To ensure fairness, we use the same GroundingDINO~\cite{liu2023grounding} model as LaMOTer to provide detection results for these trackers. Additionally, we include a comparison with TransRMOT~\cite{wu2023referring}, which is the \textit{only} open-source approach suitable for Language-Guided MOT. As in Tab.\ref{tab:baseline-comparison}, LaMOTer achieves the best performance. For instance, it achieves 48.45$\%$ in HOTA, and 47.66$\%$ in IDF1. Although LaMOTer does not achieve the highest scores on all metrics, it still attains comparable results. Meanwhile, We also find that feature-based ReID does not significantly affect the performance, \textit{e.g.}, compared to SORT, DeepSORT only achieves a +0.51$\%$ increase in HOTA and a +1.09$\%$ increase in MOTA. This is likely because descriptive language in Vision-Language MOT makes targets appear more similar, challenging feature-based ReID and limiting its performance. In addition, Tab.~\ref{tab:baseline-comparison} shows that TransRMOT performs poorly. We argue that this is because TransRMOT doesn't consider open-vocabulary contexts during its design, which leads to the model's inability to recognize unseen categories and vocabulary. These experimental results also serve as indirect evidence of LaMOT's unique open-vocabulary setting. In addition to quantitative evaluations, we also provide qualitative results of LaMOTer in Fig.~\ref{fig:results}, showcasing its performance visually.

\begin{table*}[t]
\centering
\renewcommand{\arraystretch}{1}
\tabcolsep=1.2mm
\caption{Comparison between LaMOTer and a series established trackers. Ths best and second results are highlighted in \textcolor{red}{red} and \textcolor{blue}{blue}, respectively.}
\resizebox{\textwidth}{!}{
\begin{tabular}{l|ccccccccccc}
    \toprule[1.2pt]
    \textbf{Method} & \textbf{HOTA}$\uparrow$ & \textbf{AssA}$\uparrow$ & \textbf{DetA}$\uparrow$ & \textbf{LocA}$\uparrow$ & \textbf{MOTA}$\uparrow$ & \textbf{FN}$\downarrow$ & \textbf{FP}$\downarrow$ & \textbf{IDs}$\downarrow$ & \textbf{IDR}$\uparrow$ & \textbf{IDP}$\uparrow$ & \textbf{IDF1}$\uparrow$\\
    \midrule[0.8pt]
    \textbf{SORT}~\cite{bewley2016simple} & 35.27 & 26.15 & 47.64 & 88.26 & 42.35 & 662865 & 439176 & 28298 & 31.48 & 35.54 & 33.39 \\
    \textbf{DeepSORT}~\cite{wojke2017simple} & 35.78 & 27.91 & 48.48 & 88.41 & 43.54 & 641297 & 431019 & 26712 & 34.76 & 37.67 & 35.34 \\
    \textbf{BYTETrack}~\cite{zhang2022bytetrack} & \textcolor{blue}{46.13} & 40.34 & \textcolor{blue}{50.37} & 92.09 & \textcolor{red}{46.67} & \textcolor{red}{571906} & 467891 & \textcolor{red}{10987} & \textcolor{blue}{45.98} & 49.01 & \textcolor{blue}{45.70} \\
    \textbf{StrongSORT}~\cite{du2023strongsort} & 44.38 & 38.98 & 47.86 & 90.10 & 45.17 & 591192 & \textcolor{red}{451908} & 17986 & 42.38 & 46.79 & 42.88 \\
    \textbf{MOTRv2}~\cite{zhang2023motrv2} & 45.19 & \textcolor{blue}{41.19} & 49.67 & 89.86 & 45.09 & \textcolor{blue}{581937} & 462344 & 13874 & 45.67 & \textcolor{blue}{49.10} & 45.46 \\
    \textbf{TransRMOT}~\cite{wu2023referring} & 27.74 & 21.33 & 39.56 & 80.17 & 31.67 & 782980 & 57643 & 26754 & 26.12 & 27.67 & 24.79 \\
    \midrule[0.8pt] 
    \textbf{LaMOTer(Ours)} & \textcolor{red}{48.45} & \textcolor{red}{43.92} & \textcolor{red}{53.50} & \textcolor{red}{92.91} & \textcolor{blue}{46.52} & 582036 & \textcolor{blue}{454253} & \textcolor{blue}{12355} & \textcolor{red}{46.11} & \textcolor{red}{49.32} & \textcolor{red}{47.66} \\
    \bottomrule[1.2pt]
\end{tabular}
\label{tab:baseline-comparison}
\vspace{-1.0em}
}	
\end{table*}

\begin{figure*}[!t]
 \centering
 \includegraphics[width=1.0\linewidth]{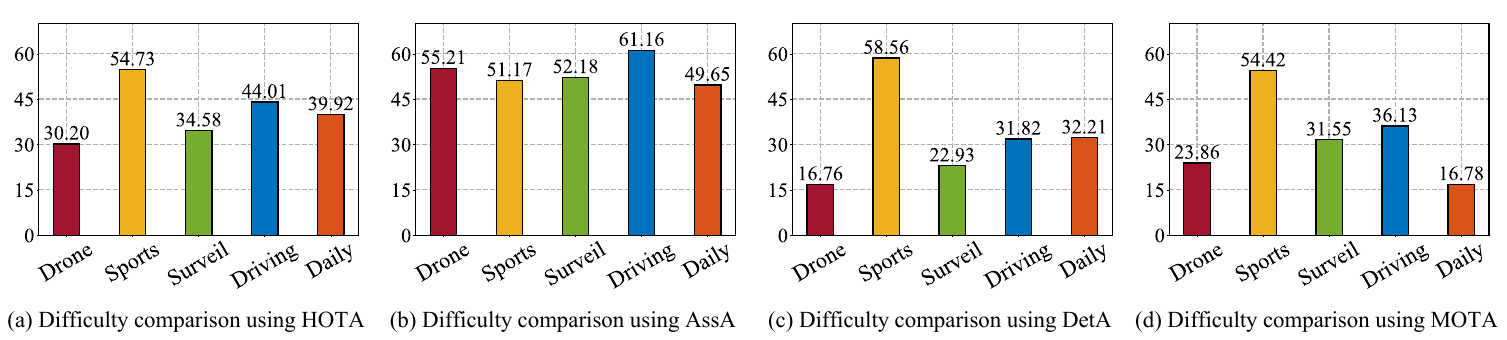}
 \caption{Difficulty comparison of different scenarios in LaMOT using different metric including HOTA (image(a)), AssA (image(b)), DetA (image(c)), and MOTA (image(d)).}
 \label{fig:difficulty}
 \vspace{-5mm}
\end{figure*}

\subsection{Difficulty Comparison of Scenarios}

To further demonstrate LaMOT, we conduct a comparison of tracking difficulty of different video scenarios. Specifically, we employ LaMOTer to evaluate the difficulty by analyzing its performance scores on subsets from various scenarios. Fig.~\ref{fig:difficulty} depicts the comparison, where larger the score is, and the less difficult the scenario is. From Fig.~\ref{fig:difficulty}, overall, the scenario of \textit{sports broadcasting} (we use sports for short in Fig.~\ref{fig:difficulty}) is the easiest to track while \textit{drone} is the most difficult based on the HOTA score (see Fig.~\ref{fig:difficulty}(a)). We believe that sports broadcasting videos, \textit{i.e.}, \textit{volleyball}, \textit{soccer}, and \textit{basketball} scenario, are relatively easy to track due to their minimal variations within each scenario. We argue that \textit{drone} videos are the hardest because its high target density and relatively low resolution, which results in difficulties for detection (see DetA score in Fig.~\ref{fig:difficulty}(c)). By conducting this difficulty analysis, we hope to guide researchers to focus more on challenging video scenarios.




\section{Conclusion}

In this paper, we propose Language-Guided MOT by unifying different Vision-Language MOT tasks. To facilitate its research, we present the large-scale benchmark LaMOT by including 1,660 sequences with 5 different scenarios, totaling 1.67M frames. To the best of our knowledge, LaMOT is the \textit{first} dataset applicable to Language-Guided MOT and is also the \textit{largest} and \textit{most challenging} dataset for Vision-Language MOT. Additionally, we propose a simple yet effective tracker, LaMOTer, and comprehensively conduct various evaluations. Through these efforts, we provide benchmarks and references to help future research understand the challenges and opportunities in Language-Guided MOT, guiding algorithm design and improvement. We also hope this paper will advance the field of Vision-Language MOT, promote the integration of computer vision and natural language processing, and lead to more intelligent and flexible tracking systems.

\end{document}